\documentclass[11pt]{article}
\usepackage[letterpaper,margin=1in]{geometry}
\usepackage{times}
\usepackage[T1]{fontenc}
\usepackage[utf8]{inputenc}
\usepackage{microtype,amsmath,amssymb,booktabs,graphicx,multirow,tabularx,array,placeins}
\usepackage[table]{xcolor}
\usepackage[round,authoryear]{natbib}
\usepackage{hyperref,url}
\definecolor{shade}{RGB}{240,245,250}
\definecolor{linkblue}{RGB}{0,64,128}
\hypersetup{colorlinks=true,linkcolor=linkblue,citecolor=linkblue,urlcolor=linkblue,
 pdftitle={Cache the Encoder Within: Compact, Reusable Memory across LLM Queries},
 pdfauthor={Hanzuo Liu, Chunyu Liu, Chaofan Lin, Alex Lamb, Mingyu Gao}}
\newcommand{\Write}{\textsc{Write}}
\newcommand{\Read}{\textsc{Read}}
\newcommand{\Select}{\textsc{Select}}
\newcolumntype{Y}{>{\raggedright\arraybackslash}X}
\title{Cache the Encoder Within:\\Compact, Reusable Memory across LLM Queries}
\author{%
Hanzuo Liu$^{1}$ \quad Chunyu Liu$^{1}$ \quad Chaofan Lin$^{2}$\\
Alex Lamb$^{1}$ \quad Mingyu Gao$^{1}$\\[0.5em]
\normalsize $^{1}$Tsinghua University (tsinghua.edu.cn)\\
\normalsize $^{2}$Tsinghua University (mail.tsinghua.edu.cn)\\[0.3em]
\normalsize \texttt{lhz24@mails.tsinghua.edu.cn}
}

\date{}
\begin{document}
\maketitle
\begin{abstract}
Repeated queries over shared documents incur redundant encoding, while
caching model states introduces persistent storage costs. Building on
CoMem's intermediate-state interface, EncBank treats a pretrained LLM's
lower layers as a reusable document encoder and compactly stores their
outputs for an adapted upper-layer reader. A self-distilled suffix adapter
is shared across storage precisions within each backbone, without
quantization-specific retraining. Across five benchmark suites on three
Qwen backbones spanning different sizes and full-attention and hybrid
architectures, 4-bit storage keeps each reported benchmark aggregate
within one score point of native-precision EncBank. In a fixed Qwen3-8B
workload, it retains 28.1\% of the native-precision persistent GPU store.
Separate native-precision controls yield a $1.40\times$ selected-pack
prefill speedup over same-evidence, same-adapter text replay, at a
3.12-point RULER accuracy cost. A native-precision Qwen3.8-27B
configuration also passes 70 of 89 Terminal-Bench 2.1 tasks. EncBank thus
combines reusable computation with compact memory, while task fidelity
and end-to-end benefits remain dependent on the workload, preparation
costs, and reuse frequency.
\end{abstract}

\section{Introduction}
Recent advances in large language models (LLMs) have enabled systems
that answer questions over long documents, retrieve information from
external knowledge collections, and draw on extended conversation histories
\citep{rag,longbench,locomo}. In these settings, a query is accompanied
by \emph{context}: source material that the model uses to construct its
answer. The same collection or history may support many different queries,
creating an opportunity to reuse the computation spent processing that material.

A straightforward approach encodes the relevant context anew for each
request before generating an answer. Retrieval reduces this cost by selecting
relevant passages, but does not by itself reuse their previous computation:
selected tokens still pass through every decoder layer. Retaining processed
representations could avoid some of this repeated work. However, those
representations must persist between requests, so their storage cost grows
with the retained collection. Efficient reuse therefore requires deciding
both \emph{what computation to retain} and \emph{how compactly to store it}.

Storing text as token IDs is compact but preserves no model computation.
Full-depth key--value (KV) caches retain more computation, but store
attention states at every layer and must handle context and position
dependencies when chunks are assembled \citep{promptcache,cacheblend,turborag}.
CoMem offers an intermediate object: one hidden state (residual) per token
at a chosen layer boundary, followed by continuation through the remaining layers
\citep{comem}. On Qwen3-8B, this costs 8 KiB/token instead of
144 KiB/token for full-depth 16-bit KV. Yet 128k tokens still require
1 GiB of residuals. Bounded retrieval limits the online working set;
it does not limit the bytes retained for a growing document collection.

EncBank extends CoMem's \Write--\Select--\Read{} interface with low-bit
persistent storage. Lower layers encode each chunk once; its states are
stored in H16, H8, or H4 format. For each query, selected states are
reconstructed and read by the upper layers. One adapted reader per backbone
serves all three precisions. Split depth $j$ controls prepaid computation;
precision $b$ controls retained bytes.

\section{Design Insights for Reusable Memory}
\label{sec:insights}
\begin{figure}[t]
\centering
\includegraphics[width=\linewidth]{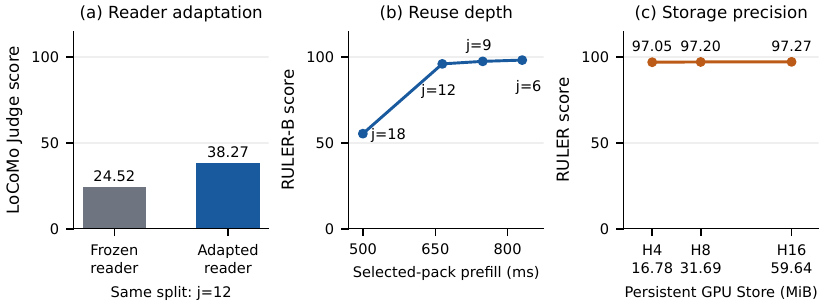}
\caption{\textbf{Reader adaptation, reuse depth, and storage precision.}
Qwen3-8B throughout. (a) Native LoCoMo full Judge, 1,986 items,
fixed $j=12$ and evidence; only suffix adaptation changes.
(b) Separately adapted splits: RULER-B on 1,500 examples versus
fixed-pack prefill from three processes.
(c) Fixed-reader FP16 H16/H8/H4: RULER means over 1,500 examples
and persistent GPU Store on six 8k documents. Quality and storage
use distinct supports. Lines connect measured settings; the depth and
precision studies are not a joint sweep.}
\label{fig:sensitivity}
\end{figure}

The diagnostics use Qwen3-8B, independently encoded 512-token chunks,
and top-12 selection. Native depth reuse (BF16) and storage precision
(FP16) are separate studies. Figure~\ref{fig:sensitivity} collects their
design evidence; prefill timing excludes Write, fetch, and decode.

\paragraph{Intermediate states benefit from reader adaptation.}
Caching after the first $j$ layers lets the remaining layers continue
processing from that boundary. However, independent chunk encoding changes
these states. At $j=12$, adapting only the upper layers raises native
LoCoMo from 24.52 to 38.27 on 1,986 items
(Figure~\ref{fig:sensitivity}a). This supports a reusable encoder with an
adapted reader; it does not establish that semantic processing is complete
at the cache boundary.

\paragraph{Deeper reuse reduces prefill but can degrade quality.}
Moving from $j=6$ to $j=12$ reduces selected-pack prefill from 830.3 to
664.4 ms, with RULER-B decreasing from 98.29 to 96.07
(Figure~\ref{fig:sensitivity}b). At $j=18$, prefill falls to 499.5 ms
but accuracy drops to 55.41. These separately adapted configurations motivate
$j=12$ as a practical compromise. Adapter sizes vary from 72.745M to
43.647M parameters, so the sweep is not a training-compute-matched
causal comparison or a held-out optimum.

\paragraph{Storage precision controls retained bytes.}
At fixed hidden width, changing $j>0$ preserves the number and width of
stored token states; deeper reuse alone does not shrink their payload.
Reducing storage precision does. With the reader fixed, H4 retains
28.1\% of H16's persistent GPU Store while RULER changes from 97.27
to 97.05 (Figure~\ref{fig:sensitivity}c). This motivates testing
precision at a fixed reader, separately from depth reuse.
The resulting design combines a chosen reuse depth, reader adaptation,
and compact states reconstructed before Read.

\section{Methodology}
\label{sec:method}
\begin{figure}[t]
\centering
\includegraphics[width=\linewidth]{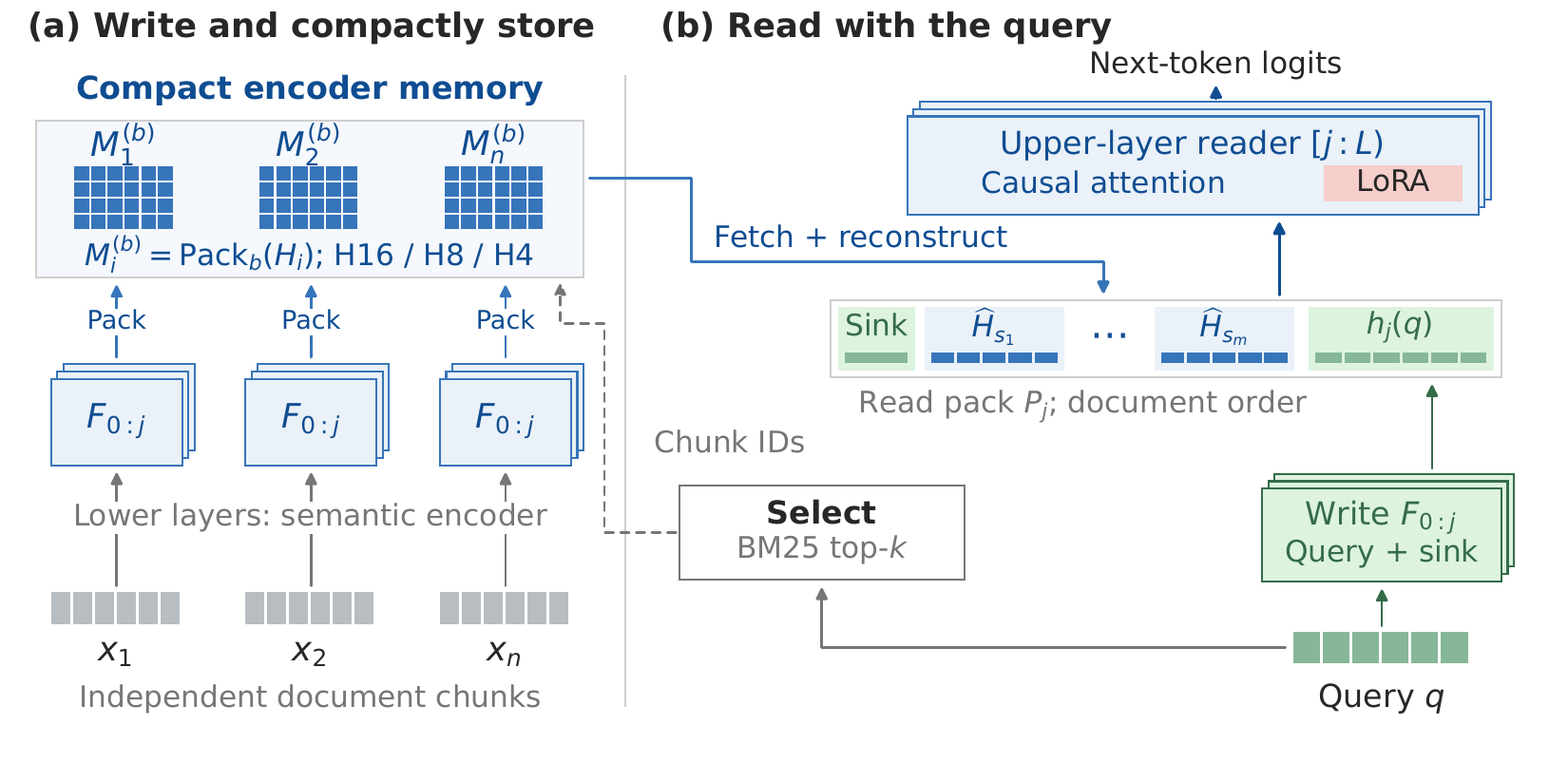}
\caption{\textbf{A persistent memory boundary inside the decoder.}
Independent lower-layer \Write{} produces document residuals, stored in
H16/H8/H4 format. Token-ID retrieval selects chunks; only their states
are reconstructed. The native sink and query join the selected states
for causal upper-layer \Read{} with suffix LoRA. The diagram describes
the attention-only Qwen3-8B implementation.}
\label{fig:architecture}
\end{figure}

Let an $L$-block decoder have hidden width $d$. The encoder $F_{0:j}$
includes embeddings and blocks $[0:j)$; $F_{j:L}$ is the reader.
A fixed source is partitioned into chunks $x_i$ of at most $c$ tokens
before queries arrive. We keep one state per token, rather than pooling
a chunk into one embedding. Figure~\ref{fig:architecture} shows the interface.

\subsection{Write once and store compactly}
Each chunk is encoded independently, with lower-layer positions restarted
at zero, and packed at precision $b\in\{16,8,4\}$:
\begin{equation}
 H_i=F_{0:j}(x_i)\in\mathbb R^{|x_i|\times d},\qquad
 M_i^{(b)}=\operatorname{Pack}_b(H_i).
 \label{eq:write}
\end{equation}
The entry retains source IDs for retrieval and a native one-token sink,
but no lower-layer document KV. H16 preserves the original BF16 or FP16
values. H8/H4 flatten each chunk into groups of $g=64$ values and use
standard affine quantization. For each group, let $a=\min_t v_t$ and
$\delta=(\max_t v_t-a)/(2^b-1)$, with $\delta=1$ for a constant group:
\begin{equation}
 u_t=\operatorname{clip}_{[0,2^b-1]}
 \left(\operatorname{round}\frac{v_t-a}{\delta}\right),\qquad
 \widehat v_t=\operatorname{cast}_{\tau}
 \left(u_t\bar\delta+\bar a\right).
 \label{eq:codec}
\end{equation}
Codes are computed in FP32. Stored BF16 metadata $\bar a,\bar\delta$
are promoted to FP32 for reconstruction, then cast to native dtype $\tau$.
H8/H4 pack one/two codes per byte and release the original tensor.
Weights, arithmetic, query/sink states, and request-local KV remain native;
H4 changes storage, not model execution.

\subsection{Retrieve, reconstruct, and read}
Iterative token-ID BM25 selects up to $k$ chunks, deduplicates them,
and restores source order. The default uses $k=12$, chunk size 512,
and three rounds of four chunks. Because selection does not use quantized
states, it is identical across matched precision arms.
For selected IDs $s_1,\ldots,s_m$, query $q$, and sink $s$, form
\begin{equation}
 P_j=[h_j(s);\widehat H_{s_1};\ldots;\widehat H_{s_m};h_j(q)],
 \qquad \widehat H_{s_r}=\operatorname{Unpack}_b(M_{s_r}^{(b)}).
 \label{eq:read}
\end{equation}
The sink and query are encoded independently. The suffix applies fresh
contiguous positions and causal attention: later chunks see earlier chunks,
and query tokens see all selected chunks. Every generated token still
traverses all $L$ blocks. In the attention-only implementation, lower-layer
KV contains the query/generated prefix; upper-layer KV additionally contains
the selected pack. Their position histories advance separately.
Reconstructed states and generation caches are request-local; stored document
entries persist. Hybrid backbones retain the same residual interface with
architecture-specific attention/recurrent-state handling.

Independent encoding changes the states consumed by the suffix.
Following CoMem, self-distillation trains LoRA on suffix attention and MLP
projections while freezing the encoder and backbone \citep{lora,distillation}.
The adapter-disabled teacher continuously processes ordered PG-19 text
\citep{pg19}; the student separately encodes chunks, sink, and query.
At query position $t$, teacher top-64 indices define support $S_t$;
teacher $p_t$ and student $q_t$ are renormalized on this support:
\begin{equation}
 \mathcal L=\frac{1}{|\mathcal Q|}\sum_{t\in\mathcal Q}
 [0.6\,D_{\mathrm{KL}}(p_t\Vert q_t)+0.4\,D_{\mathrm{KL}}(q_t\Vert p_t)].
 \label{eq:distill}
\end{equation}
Training uses native states without retrieval, hidden-state quantization,
or benchmark supervision. H16/H8/H4 share the resulting reader within
each backbone. The conditional objective does not constrain total student
mass on $S_t$; we test full-vocabulary behavior separately.

\subsection{Controlled endpoints and cost accounting}
At $j=0$, raw IDs for the same selected chunks are replayed through all
layers. Holding evidence, order, sink, adapter, and generation fixed isolates
the effect of independent residual reuse. A continuous-prefix oracle
instead computes $h_j$ on the assembled pack afresh; it tests continuation
correctness but loses cross-query reuse. H8/H4 versus H16 is a second
control, varying only storage precision at fixed depth and reader.

For $N$ stored tokens, the document-state payload is
\begin{equation}
 S_{16}=2Nd,\qquad S_b\simeq Nd\left(\frac b8+\frac4g\right),\quad b=8,4.
 \label{eq:storage}
\end{equation}
At $g=64$, H8/H4 retain 53.125\%/28.125\% of H16 payload.
For Qwen3-8B ($L=36,d=4096,n_{\rm kv}=8,d_{\rm head}=128$),
H16/H4 require 8/2.25 KiB per token, versus
$4Ln_{\rm kv}d_{\rm head}=144$ KiB for full-depth 16-bit KV.
These exclude weights, retrieval metadata, and request workspace;
the KV formula does not describe hybrid recurrent-state storage.
The selected pack is bounded by $1+kc+|q|$ positions, but Write,
indexing, lookup, and total storage still grow with the collection.

\section{Experimental Design}
\label{sec:setup}
We separate \emph{native-reuse controls} from \emph{storage-precision
comparisons}. The former use BF16 Qwen3-8B and include matched replay,
depth, adaptation, context, and serving tests. The latter use a custom
rank-32, $\alpha=32$, 4,000-step reader, with a complete FP16 Qwen3-8B
suite at $j=12$. Its document boundaries are fixed before the query;
some native quality harnesses instead use the last formatted-prompt chunk
as the query. Checkpoint identity is not established across these studies:
we neither add their quality gaps nor transfer speedups to H4.

Reported extensions use Qwen3.5-9B and Qwen3.8-27B hybrid backbones at
$j=6,21$, each with its own reader \citep{qwen3,qwen35config,qwen38config}.
They use native chat templates with thinking disabled; Qwen3-8B uses
plain-text greedy decoding. Hybrid low-bit results are supplied aggregates,
with less complete checkpoint/judge binding than the 8B study.

Five benchmarks cover retrieval and reasoning: RULER (three tasks,
five lengths, 1,500 examples), LongEval (five lengths, 500),
LongBench QA6 (six datasets, 1,150), BABILong (three tasks, seven lengths,
2,100), and LoCoMo \citep{ruler,longchat,longbench,babilong,locomo}.
We retain native scorers and macro weighting; LoCoMo uses item weighting.
The 8B precision judge covers 1,540 answerable items; native-reuse full
Judge includes 446 additional abstention items. These scores are not pooled.
Systems use Qwen3-8B and report persistent Store, peak memory, Write,
TTFT, decode, and preparation-inclusive E2E separately. Core evidence
and counterexamples follow; reproduction details are in Appendix~\ref{app:protocol}.

We evaluate the 89-task Terminal-Bench 2.1 agent benchmark
\citep{terminalbench21} with Qwen3.8-27B. EncBank-H16 divides prior chat
and tool history into 512-token chunks, reuses their H21 states, and uses
token-ID BM25 to select up to 12 relevant chunks ($k=12$) for the adapted
suffix reader alongside the current exchange. Dense rereads the full
history; task verifiers score outcomes. Both configurations cover all 89
tasks; adapter and serving paths differ.

Code repository: \url{https://anonymous.4open.science/r/EncBank-1E15/}.

\section{Results}
\subsection{Low-bit memory preserves benchmark aggregates, not every answer}
\label{sec:quality}
\begin{table}[t]
\centering\small
\setlength{\tabcolsep}{4pt}
\caption{\textbf{Quality across storage precisions and backbones.}
Scores are 0--100. Within each H16/H8/H4 group, the reader is fixed;
other rows are whole-method references. 8B values follow the complete
FP16 precision study, not the separate native-reuse cohort.
Hybrid values are reported aggregates ($\dagger$); see the scope below.}
\label{tab:quality}
\begin{tabular*}{\linewidth}{@{\extracolsep{\fill}}lrrrrr@{}}
\toprule
Method & RULER & LongEval & QA6 F1 & BABILong & LoCoMo\\
\midrule
\multicolumn{6}{l}{\textit{Qwen3-8B; full attention; $j=12$; FP16 precision study}}\\
Dense, full source & 51.91 & 31.6 & 8.97 & 58.52 & 32.40\\
KIVI2 & 42.43 & 9.0 & 9.37 & 53.38 & 29.81\\
KIVI4 & 50.64 & 31.2 & 8.89 & 58.90 & 32.60\\
Residual, no LoRA & 16.01 & 0.0 & 9.98 & 38.38 & 25.45\\
EncBank-H16 & 97.27 & 63.0 & 12.05 & 58.05 & 44.03\\
EncBank-H8 & 97.20 & 62.6 & 12.11 & 58.14 & 44.09\\
\rowcolor{shade}EncBank-H4 & 97.05 & 63.6 & 11.96 & 58.95 & 43.57\\
\midrule
\multicolumn{6}{l}{\textit{Qwen3.5-9B; hybrid; $j=6$; native chat}$^{\dagger}$}\\
Dense, full source & 85.87 & 99.80 & 51.98 & 73.57 & 48.59\\
Residual, no LoRA & 61.08 & 73.60 & 43.82 & 56.33 & 38.57\\
EncBank-H16 & 75.53 & 97.20 & 46.52 & 63.62 & 46.58\\
EncBank-H8 & 75.25 & 97.20 & 46.75 & 64.05 & 46.48\\
\rowcolor{shade}EncBank-H4 & 76.47 & 97.20 & 46.19 & 64.00 & 46.78\\
\midrule
\multicolumn{6}{l}{\textit{Qwen3.8-27B; hybrid; $j=21$; native chat}$^{\dagger}$}\\
Dense, full source & 100.00 & 100.00 & 56.36 & 69.14 & 55.09\\
Residual, no LoRA & 96.75 & 85.60 & 48.20 & 66.81 & 45.97\\
EncBank-H16 & 99.39 & 97.20 & 50.91 & 73.67 & 50.70\\
EncBank-H8 & 99.39 & 97.20 & 50.96 & 73.38 & 50.50\\
\rowcolor{shade}EncBank-H4 & 99.40 & 97.20 & 51.08 & 72.81 & 50.30\\
\bottomrule
\end{tabular*}
\par\smallskip
\begin{minipage}{\linewidth}\footnotesize
RULER/LongEval cover 8k--128k; 8B full-source 64k/128k are unextended
stress tests. QA6 averages six datasets; BABILong averages 21 cells.
8B LoCoMo uses 1,540 category-1--4 judgments. $\dagger$ Hybrid H16
protocols use 1,986 items; low-bit judge/checkpoint bindings are not
independently supplied. Hybrid contrasts remain descriptive, not verified
paired or cross-model rankings. No significance is inferred from maxima.
\end{minipage}
\end{table}

Table~\ref{tab:quality} isolates precision within each adapted-reader group.
On Qwen3-8B, H4 changes RULER by $-0.22$, LongEval by $+0.60$,
LongBench by $-0.09$, BABILong by $+0.90$, and LoCoMo by $-0.46$
points relative to H16. The largest absolute H4 change is 0.94 on the
9B summary and 0.86 on 27B. Thus all supplied five-benchmark aggregates
remain within one point, without a separately trained quantized reader.
This is a descriptive observation, not a non-inferiority test.

Full-source references do not isolate quantization. In the 8B suite, Dense and KIVI have
LoRA disabled and see the full source; 64k/128k exceed the unextended
40,960-position window. Their failures there do not establish superiority
over a length-extended full-context model. Within the separate native
YaRN control, full-context replay reaches 100 on 128k single-needle,
versus 98 for residual reuse. On the hybrid summaries, full-source
recomputation is stronger on most tasks. The evidence supports compact usable
memory, not a universal answer-quality advantage.

Aggregate proximity hides variation. Across 8k/16k/32k/64k/128k,
the H4-minus-H16 LongEval differences are $(4,4,-6,-3,4)$ points.
It loses 0.59 F1 on MuSiQue, whose
unadjusted paired document-bootstrap interval is $[-1.20,-0.03]$.
All LongEval precision outputs reach the 16-token cap. In the separate
BF16 single-key/8k cohort, H16/H8/H4 each score 99, but complete
sequences match H16 on only 68/100 H8 and 34/100 H4 examples.

\subsection{Matched replay separates saved computation from lost fidelity}
\label{sec:depth}
\begin{table}[t]
\centering\small
\caption{\textbf{Matched native reuse measures a fidelity cost.}
RULER-B uses 1,500 paired examples; MK is the 200-example 8k/16k
multikey subset. Prefill uses a separate fixed pack, three processes,
three warmups and 20 reads/process; it excludes selection, Write,
fetch and decode. NR: latency not reported for the per-query oracle.}
\label{tab:replay}
\begin{tabular*}{\linewidth}{@{\extracolsep{\fill}}lrrrc@{}}
\toprule
Path & RULER-B & MK & Prefill (ms) & Reusable\\
\midrule
Same-adapter replay, $j=0$ & 99.19 & 100.0 & 931.9 & No\\
Chunk-local H16, $j=12$ & 96.07 & 92.5 & 664.4 & Yes\\
Continuous-prefix oracle & 99.19 & 100.0 & NR & No\\
\bottomrule
\end{tabular*}
\end{table}

The native control (Table~\ref{tab:replay}) fixes checkpoint, all 168
suffix-LoRA modules, evidence IDs/order, examples, and decoding.
On 1,500 paired RULER-B examples, $j=12$ reuse loses 3.12 points
(95\% interval $[2.36,3.93]$). Separately, prefill on a fixed 6.5k-token
pack falls from 931.9 to 664.4 ms ($1.403\times$; three-process ratios
1.402--1.404), excluding selection, Write, fetch, and decode.
Including measured decode reduces the ratio to $1.07$--$1.09\times$.

Fresh continuous-prefix states recover replay exactly on the reported
examples, validating continuation but requiring per-query recomputation.
Native LongEval replay/reuse scores are 97.2/69.0; the later paired
LongBench comparison gives 12.18/12.01 F1. LongBench replay is adapter-free,
so it does not provide the same-adapter attribution of RULER.

\subsection{Adaptation and evidence interaction matter; local repair is limited}
\label{sec:mechanisms}
\begin{table}[t]
\centering\small
\caption{\textbf{Context interaction and local repair, with separate cohorts.}
(a) Multikey full/block-diagonal attention, same evidence,
50 examples/length; query sees all chunks.
(b) Left-overlap Write at fixed stored shapes: MK has 200 examples;
independent LongEval (LE) has 300, Qasper 200. Scores are 0--100.}
\label{tab:mechanisms}
\begin{tabular*}{\linewidth}{@{\extracolsep{\fill}}llrrr@{}}
\toprule
Panel & Configuration & \multicolumn{3}{c}{Measured outcomes}\\
\midrule
\multirow{3}{*}{(a) Interaction} & & MK/8k & MK/16k & \\
 & Full causal & 96 & 94 & \\
 & Block diagonal & 60 & 32 & \\
\midrule
\multirow{3}{*}{(b) Write context} & & MK & LE & Qasper F1\\
 & No overlap & 92.5 & 72.67 & 11.37\\
 & 32-token overlap & 98.5 & 62.33 & 10.76\\
\bottomrule
\end{tabular*}
\end{table}

Beyond the LoCoMo gain in Section~\ref{sec:insights}, adaptation
also improves predictive fidelity on 50 held-out PG-19 books
(100 windows, 51,200 query positions): full-vocabulary forward KL
falls from .442 to .304 and next-token NLL from 2.680 to 2.579;
the paired book-bootstrap changes are $-.138\;[-.147,-.131]$ and
$-.101\;[-.111,-.092]$. Teacher top-64 support retains 94.68\%
probability mass on average. Improvement therefore extends beyond the
support-normalized objective, without making it full-vocabulary training.

Blocking evidence-to-evidence attention reduces multikey accuracy from
96/94 to 60/32 at 8k/16k (Table~\ref{tab:mechanisms}), although queries
see every selected chunk in both arms.

On a paired 200-example multikey diagnostic, full-document lower-layer
context raises 92.5 to 100, while changing only upper-layer positions
lowers it to 88. The full-context arm also changes query states and
lower-layer decode visibility, so it does not isolate document states alone.
A 32-token left overlap raises multikey to 98.5 with unchanged stored
shapes. Yet on 300 independent LongEval examples, the same overlap
reduces accuracy by 10.33 points ($[-16.33,-4.33]$), and it does not
improve Qasper. Its measured Write cost increases 17.9--19.4\%.
We retain disjoint chunks as the default and do not generalize the
synthetic repair result to natural QA.

\subsection{Single-residual and full-depth KV objects trade different costs}
\label{sec:kv}
\begin{table}[t]
\centering\small
\caption{\textbf{Matched evidence exposes the residual--KV trade-off.}
(a) Quality on 500 examples: LongEval at 8k/16k/32k (100 each),
Qasper (200). The shared adapter was trained for EncBank.
(b) RTX 5090 costs on five prespecified LongEval/32k examples,
shared adapter on, three processes and three repetitions/example.
Preparation builds the whole-document pinned-CPU store; E2E includes
128 output tokens. Chunk-KV is a synchronous CacheBlend-style control.}
\label{tab:kv}
\begin{tabular*}{\linewidth}{@{\extracolsep{\fill}}llrrr@{}}
\toprule
(a) Method & Adapter & LongEval & Qasper F1 & KiB/token\\
\midrule
Replay & off & 84.33 & 6.98 & .008\\
Chunk-KV & off & 73.00 & 4.88 & 144\\
Residual H16 & off & 0.00 & 10.45 & 8\\
Replay & on & 77.67 & 4.65 & .008\\
Chunk-KV & on & 55.67 & 3.68 & 144\\
EncBank-H16 & on & 72.67 & 11.37 & 8\\
\bottomrule
\end{tabular*}
\smallskip
\begin{tabular*}{\linewidth}{@{\extracolsep{\fill}}lrrrr@{}}
\toprule
(b) Method & Prep. (s) & TTFT (ms) & E2E$_{128}$ (s) & Store (MiB)\\
\midrule
Replay & .0011 & 803.7 & 6.400 & .25\\
Chunk-KV & 5.7085 & 399.9 & 11.694 & 4608\\
EncBank-H16 & 1.3192 & 613.2 & 7.012 & 256\\
\bottomrule
\end{tabular*}
\end{table}

A same-evidence, same-backbone control compares replay, EncBank-H16,
and a synchronous CacheBlend-style chunk-KV implementation
\citep{cacheblend}. It caches all 36 layers, reindexes RoPE,
recomputes two bootstrap layers, and repairs a fixed 15\% context-token
set above them. It is not the native cache manager or asynchronous system.
All paths are tested with the EncBank suffix adapter enabled and disabled;
no KV-specific adapter is trained.

Adaptation helps residual reuse but hurts the other interfaces
(Table~\ref{tab:kv}a). Its 17-point LongEval advantage over adapted
chunk-KV disappears against unadapted chunk-KV (72.67 versus 73.00).
Qasper favors adapted residuals, though all scores remain low under
capped plain-text generation. Thus the shared-adapter comparison does
not establish an intrinsic residual-over-KV advantage.

On five prespecified 32k examples (Table~\ref{tab:kv}b), chunk-KV
has lower TTFT but 18 times the payload and substantially more
preparation. Residual reuse reduces cold E2E relative to chunk-KV,
yet remains slower than same-adapter replay when Write is charged.
This 128-output-token cost subset is separate from the 500-example quality study.

\subsection{Low-bit storage reduces residency cost, not proportional request cost}
\label{sec:systems}
\begin{table}[t]
\centering\small
\setlength{\tabcolsep}{4pt}
\caption{\textbf{Persistent storage savings versus request costs.}
RTX 5090, FP16, six RULER/8k documents, 18 fresh Write/Read pairs,
32 fixed output tokens. Only H variants share the reader.
Warm TTFT excludes document Write. Store is persistent GPU MiB;
peak is sampled whole-device GiB, including weights. Write
times are component means; decode throughput is pooled.}
\label{tab:systems}
\begin{tabular*}{\linewidth}{@{\extracolsep{\fill}}lrrrrr@{}}
\toprule
Method & Store & Write (ms) & Warm TTFT (ms) & Tokens/s & Peak\\
\midrule
Dense & 1073.51 & 710.68 & 654.71 & 44.96 & 21.223\\
KIVI2 & 210.83 & 1127.73 & 1485.31 & 20.19 & 20.604\\
KIVI4 & 343.55 & 1128.38 & 1486.14 & 19.99 & 20.603\\
EncBank-H16 & 59.64 & 265.62 & 1382.98 & 33.56 & 20.648\\
EncBank-H8 & 31.69 & 274.12 & 1392.93 & 32.95 & 20.627\\
\rowcolor{shade}EncBank-H4 & 16.78 & 271.76 & 1378.10 & 32.97 & 20.617\\
\bottomrule
\end{tabular*}
\end{table}

On a fixed RTX 5090 FP16 workload, H8/H4 retain 53.1\%/28.1\%
of H16 persistent GPU Store (Table~\ref{tab:systems}).
Warm TTFT stays near 1.38 s and sampled peaks near 20.6 GiB;
decode is slightly slower. Weights, reconstruction buffers, and request KV
remain: 71.9\% Store savings do not proportionally reduce request costs.
Dense is faster in this warm workload; KIVI has slightly lower sampled
peaks. Different adapters and execution paths preclude attributing these
whole-method differences solely to the persistent representation.

This fixed 32-output-token fresh-entry test is not a repeated-query
amortization measurement. Component timing, sampled peaks, and Store
accounting are specified in Appendix~\ref{app:protocol}.

\begin{figure}[t]
\centering
\includegraphics[width=\linewidth]{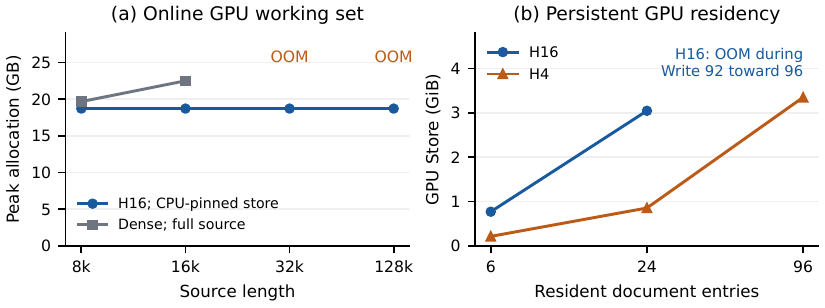}
\caption{\textbf{Bounded online allocation and residency.}
(a) Native H16/CPU-pinned online TTFT-phase peak on RTX 5090; full-source
Dense fails at 32k/128k under a 28 GB allocator cap. (b) Separate FP16
GPU-resident test with distinct documents: H16 fails at Write 92,
H4 completes 96. Levels are measured, not maximum-capacity estimates.}
\label{fig:scale}
\end{figure}

Without eviction/offload, H4 completes 96 distinct MuSiQue entries using
3,431.26 MiB of GPU Store; H16 fails on Write 92
(Figure~\ref{fig:scale}b). H4 warm TTFT at 6/24/96 entries is
1877.57/1847.44/1857.58 ms on identical queries to six known documents.
This tests residency, not collection-wide retrieval; native F1 is 10.86.

\subsection{Scaling and query reuse determine whether preparation pays off}
\label{sec:amortization}
\enlargethispage{\baselineskip}
With top-12 selection and CPU-pinned H16, online peak allocation stays
near 18.7 GB from 8k to 128k (Figure~\ref{fig:scale}a). TTFT rises from
652.6 to 768.6 ms; preparation-inclusive 128-token E2E rises from 6.332
to 11.427 s. Dense fails at 32k/128k under a 28 GB allocator cap; H16
uses 26 GB. These unscored measurements do not establish quality beyond
native positions.
\newpage

Let $W_b,I_0$ denote cached/raw-token preparation and $T_b,T_0$ per-query
costs including fetch, Read, and decode. With common selection cancelling,
\begin{equation}
 Q_b^\star=\frac{W_b-I_0}{T_0-T_b},\qquad T_0>T_b.
 \label{eq:breakeven}
\end{equation}
\begin{table}[t]
\centering\small
\caption{\textbf{Native H16 preparation payback depends on reuse.}
$Q^\star$ in queries from measured Write, fetch, Read and decode;
$G$ is output tokens. Three process medians; stable stores and cache hits.
Round up for an integer threshold. $\infty$ means nonpositive per-query
savings. This grid is not an H4 amortization measurement.}
\label{tab:amortization}
\begin{tabular*}{\linewidth}{@{\extracolsep{\fill}}lrrrr@{}}
\toprule
Source / store & $G=1$ & $G=32$ & $G=128$ & $G=512$\\
\midrule
32k / CPU pinned & 8.9 & 9.2 & 10.9 & 94.0\\
32k / GPU resident & 8.4 & 7.7 & 5.5 & $\infty$\\
128k / CPU pinned & 27.6 & 29.7 & 37.4 & 520.1\\
128k / GPU resident & 25.8 & 26.8 & 27.2 & 164.2\\
1M / CPU pinned & 198.2 & 266.7 & 421.3 & 302.6\\
1M / GPU resident & 180.2 & 183.0 & 190.6 & 574.7\\
\bottomrule
\end{tabular*}
\end{table}

The native grid (Table~\ref{tab:amortization}) gives 9--11 queries for
32k CPU-pinned sources and outputs up to 128 tokens. Longer sources
increase preparation; longer outputs can erase savings. At 32k/GPU
with 512 output tokens, no finite crossover occurs. These medians assume
stable stores and cache hits; they do not transfer to H8/H4.

Bounded reading also bounds evidence coverage: the retained store-scale
diagnostic reads about 6.2--6.5k tokens from 128k--4M stored tokens but
does not solve global aggregation. At top-12, recall for 16/24/32 required
chunks is .75/.50/.375, and a common-word-frequency task scores zero.

\subsection{Agentic terminal tasks: full-suite verifier outcomes}
\label{sec:terminalbench}
\begin{table}[t]
\centering\small
\setlength{\tabcolsep}{4pt}
\caption{\textbf{Terminal-Bench 2.1 verifier results on all 89 tasks.}
Each method retains one verified outcome per task; adapters and serving
backends differ.}
\label{tab:terminalbench}
\begin{tabular*}{\linewidth}{@{\extracolsep{\fill}}lrr@{}}
\toprule
Method & Pass / tasks & Success (\%)\\
\midrule
Dense, full context & 61/89 & 68.54\\
EncBank-H16, $k=12$ & 70/89 & 78.65\\
\bottomrule
\end{tabular*}
\end{table}

EncBank-H16 at $k=12$ passes 70/89 tasks (78.65\%), versus Dense's
61/89 (68.54\%). Different adapters and serving backends make this a
descriptive comparison, not an isolated memory-reuse effect or official
leaderboard result.

\subsection{Robustness checks expose the limits of the operating point}
\label{sec:robustness}
\paragraph{Can saved compute buy enough evidence?}
A disjoint latency calibration permits ten replay chunks versus twelve
for H16. Across nine 100-example cells, BM25 replay scores 64.78 versus
53.22 for H16 ($-11.56$ points, 95\% interval $[-18.67,-5.11]$).
Frozen-BGE replay narrows this to $-1.00\;[-10.67,8.33]$.
The one-conversation LoCoMo slice and missing per-cell quality latencies
preclude a task-wide equal-latency claim.

\paragraph{Corpus overlap and prompt sensitivity.}
After 13-gram clean filtering, 1,053/1,150 LongBench items remain;
adapted/no-adapter/replay macros are 12.05/9.90/12.21. Adaptation gains
2.15 points ($[1.57,2.72]$), but its replay gap is unresolved.
Chat formatting reverses native LoCoMo EncBank/full-source ordering
(38.27/34.59 to 37.76/38.22); a 200-item judge audit has only .81
agreement. Natural-task conclusions remain protocol-specific.

\section{Related Work}
\paragraph{Reusable representations.}
ReadOnce and Embedding Recycling reuse compressed text or activations
\citep{readonce,embeddingrecycling}. LLoCO adds offline compression and
LoRA; LongMem, XC-Cache, and ILRe use side memory, cross-attention, or
intermediate keys \citep{lloco,longmem,xccache,ilre}. CoMem retrieves
independently encoded residual chunks for suffix continuation;
EncBank adds low-bit persistent storage \citep{comem}.

\paragraph{KV reuse and restoration.}
Prompt Cache, TurboRAG, and CacheBlend organize or repair reusable KV
\citep{promptcache,turborag,cacheblend}; HCache and KV-Direct reconstruct
it from intermediate states \citep{hcache,kvdirect}. Cartridges, SemPIC,
and KV Packet learn cache representations \citep{cartridges,sempic,kvpacket}.
EncBank instead stores one split-layer state and adapts once.

\paragraph{Compression target.}
KIVI and KVQuant quantize attention KV, CacheGen compresses it, and
SmoothQuant targets weight--activation computation
\citep{kivi,kvquant,cachegen,smoothquant}. EncBank instead quantizes
stored residuals and measures storage, quality, and request cost separately.

\section{Limitations and Conclusion}
Depth and precision are not jointly optimized. Controls center on
Qwen3-8B; hybrid low-bit provenance and a paired replay-to-H4 estimate
are lacking. Eviction, throughput, tail latency, dynamic updates,
long-form and non-English generation remain untested. Persistent states
require source-equivalent access and deletion; low-bit storage offers no
privacy guarantee. EncBank reduces persistent storage while retaining
useful benchmark quality, with payback depending on task and reuse.

\bibliographystyle{plainnat}
\bibliography{refs}

@misc{comem,
      title={Understanding Is Done Early: A Depth Division of Labor in Large Language Models and Its Use for Unbounded-Context Memory}, 
      author={Hanzuo Liu and Xuan Qi and Chunyu Liu and Haotian Zhong and Yulong Wang and Rayying and Key and Alex Lamb and Mingyu Gao},
      year={2026},
      eprint={2607.28263},
      archivePrefix={arXiv},
      primaryClass={cs.CL},
      url={https://arxiv.org/abs/2607.28263}, 
}

@inproceedings{babilong,
  title={{BABILong}: Testing the Limits of {LLMs} with Long Context Reasoning-in-a-Haystack},
  author={Kuratov, Yuri and Bulatov, Aydar and Anokhin, Petr and Rodkin, Ivan and Sorokin, Dmitry and Sorokin, Artyom and Burtsev, Mikhail},
  booktitle={Advances in Neural Information Processing Systems},
  year={2024},
  url={https://proceedings.neurips.cc/paper_files/paper/2024/hash/c0d62e70dbc659cc9bd44cbcf1cb652f-Abstract-Datasets_and_Benchmarks_Track.html},
  volume={37},
  pages={106519--106554},
}

@inproceedings{cacheblend,
  title={{CacheBlend}: Fast Large Language Model Serving for {RAG} with Cached Knowledge Fusion},
  author={Yao, Jiayi and Li, Hanchen and Liu, Yuhan and Ray, Siddhant and Cheng, Yihua and Zhang, Qizheng and Du, Kuntai and Lu, Shan and Jiang, Junchen},
  year={2025},
  booktitle={Proceedings of the Twentieth European Conference on Computer Systems},
  pages={94--109},
  url={https://arxiv.org/abs/2405.16444},
  doi={10.1145/3689031.3696098},
}

@inproceedings{cartridges,
  title={Cartridges: Lightweight and General-Purpose Long Context Representations via Self-Study},
  author={Eyuboglu, Sabri and Ehrlich, Ryan and Arora, Simran and Guha, Neel and Zinsley, Dylan and Liu, Emily and Rudra, Atri and Zou, James Y and Mirhoseini, Azalia and R{\'e}, Christopher},
  year={2026},
  booktitle={International Conference on Learning Representations},
  url={https://proceedings.iclr.cc/paper_files/paper/2026/hash/4681359a7b1e94571598ad1adda35e6e-Abstract-Conference.html},
}

@article{distillation,
  title={Distilling the Knowledge in a Neural Network},
  author={Hinton, Geoffrey and Vinyals, Oriol and Dean, Jeff},
  journal={arXiv preprint arXiv:1503.02531},
  year={2015},
  url={https://arxiv.org/abs/1503.02531},
}

@inproceedings{embeddingrecycling,
  title={Embedding Recycling for Language Models},
  author={Saad-Falcon, Jon and Singh, Amanpreet and Soldaini, Luca and D'Arcy, Mike and Cohan, Arman and Downey, Doug},
  year={2023},
  booktitle={Findings of the Association for Computational Linguistics: EACL 2023},
  url={https://aclanthology.org/2023.findings-eacl.145/},
  pages={1933--1953},
  doi={10.18653/v1/2023.findings-eacl.145},
}

@inproceedings{hcache,
  title={Fast State Restoration in {LLM} Serving with {HCache}},
  author={Gao, Shiwei and Chen, Youmin and Shu, Jiwu},
  year={2025},
  booktitle={Proceedings of the Twentieth European Conference on Computer Systems},
  pages={128--143},
  url={https://arxiv.org/abs/2410.05004},
  doi={10.1145/3689031.3696072},
}

@article{ilre,
  title={{ILRe}: Intermediate Layer Retrieval for Context Compression in Causal Language Models},
  author={Liang, Manlai and Liu, Mandi and Ji, Jiangzhou and Li, Huaijun and Yang, Haobo and He, Yaohan and Li, Jinlong},
  year={2025},
  journal={arXiv preprint arXiv:2508.17892},
  url={https://arxiv.org/abs/2508.17892},
}

@article{kvdirect,
  title={The Residual Stream Is All You Need: On the Redundancy of the {KV} Cache in Transformer Inference},
  author={Qasim, Kaleem Ullah and Zhang, Jiashu and Shaheen, Muhammad Kafeel and Alharith, Razan and Zhang, Heying},
  journal={arXiv preprint arXiv:2603.19664},
  year={2026},
  url={https://arxiv.org/abs/2603.19664},
}

@article{kvpacket,
  title={{KV Packet}: Recomputation-Free Context-Independent {KV} Caching for {LLMs}},
  author={Chen, Chuangtao and Zhang, Grace Li and Yin, Xunzhao and Zhuo, Cheng and Li, Bing and Schlichtmann, Ulf},
  year={2026},
  journal={arXiv preprint arXiv:2604.13226},
  url={https://arxiv.org/abs/2604.13226},
}

@inproceedings{lloco,
  title={{LLoCO}: Learning Long Contexts Offline},
  author={Sijun Tan and Xiuyu Li and Shishir G Patil and Ziyang Wu and Tianjun Zhang and Kurt Keutzer and Joseph E. Gonzalez and Raluca Ada Popa},
  year={2024},
  url={https://aclanthology.org/2024.emnlp-main.975/},
  booktitle={Proceedings of the 2024 Conference on Empirical Methods in Natural Language Processing},
  pages={17605--17621},
  doi={10.18653/v1/2024.emnlp-main.975},
}

@inproceedings{locomo,
  title={Evaluating Very Long-Term Conversational Memory of {LLM} Agents},
  author={Maharana, Adyasha and Lee, Dong-Ho and Tulyakov, Sergey and Bansal, Mohit and Barbieri, Francesco and Fang, Yuwei},
  year={2024},
  booktitle={Proceedings of the 62nd Annual Meeting of the Association for Computational Linguistics (Volume 1: Long Papers)},
  pages={13851--13870},
  url={https://aclanthology.org/2024.acl-long.747/},
  doi={10.18653/v1/2024.acl-long.747},
}

@inproceedings{longbench,
  title={{LongBench}: A Bilingual, Multitask Benchmark for Long Context Understanding},
  author={Bai, Yushi and Lv, Xin and Zhang, Jiajie and Lyu, Hongchang and Tang, Jiankai and Huang, Zhidian and Du, Zhengxiao and Liu, Xiao and Zeng, Aohan and Hou, Lei and Dong, Yuxiao and Tang, Jie and Li, Juanzi},
  year={2024},
  booktitle={Proceedings of the 62nd Annual Meeting of the Association for Computational Linguistics (Volume 1: Long Papers)},
  pages={3119--3137},
  url={https://aclanthology.org/2024.acl-long.172/},
  doi={10.18653/v1/2024.acl-long.172},
}

@misc{longchat,
  title={How Long Can Open-Source {LLMs} Truly Promise on Context Length?},
  author={Li, Dacheng and Shao, Rulin and Xie, Anze and Sheng, Ying and Zheng, Lianmin and Gonzalez, Joseph E. and Stoica, Ion and Ma, Xuezhe and Zhang, Hao},
  howpublished={LMSYS Org Blog},
  year={2023},
  url={https://lmsys.org/blog/2023-06-29-longchat/}
}

@inproceedings{longmem,
  title={Augmenting Language Models with Long-Term Memory},
  author={Wang, Weizhi and Dong, Li and Cheng, Hao and Liu, Xiaodong and Yan, Xifeng and Gao, Jianfeng and Wei, Furu},
  booktitle={Advances in Neural Information Processing Systems},
  year={2023},
  url={https://proceedings.neurips.cc/paper_files/paper/2023/hash/ebd82705f44793b6f9ade5a669d0f0bf-Abstract-Conference.html},
  volume={36},
  pages={74530--74543},
}

@inproceedings{lora,
  title={{LoRA}: Low-Rank Adaptation of Large Language Models},
  author={Hu, Edward J. and Shen, Yelong and Wallis, Phillip and Allen-Zhu, Zeyuan and Li, Yuanzhi and Wang, Shean and Wang, Lu and Chen, Weizhu},
  booktitle={International Conference on Learning Representations},
  year={2022},
  url={https://openreview.net/forum?id=nZeVKeeFYf9},
}

@article{pg19,
  title={Compressive Transformers for Long-Range Sequence Modelling},
  author={Rae, Jack W. and Potapenko, Anna and Jayakumar, Siddhant M. and Hillier, Chloe and Lillicrap, Timothy P.},
  journal={arXiv preprint arXiv:1911.05507},
  year={2019},
  url={https://arxiv.org/abs/1911.05507},
}

@inproceedings{promptcache,
  title={Prompt Cache: Modular Attention Reuse for Low-Latency Inference},
  author={Gim, In and Chen, Guojun and Lee, Seung-seob and Sarda, Nikhil and Khandelwal, Anurag and Zhong, Lin},
  year={2024},
  booktitle={Proceedings of Machine Learning and Systems},
  volume={6},
  pages={325--338},
  url={https://proceedings.mlsys.org/paper_files/paper/2024/hash/a66caa1703fe34705a4368c3014c1966-Abstract-Conference.html},
}

@article{qwen3,
  title={{Qwen3} Technical Report},
  author={Yang, An and others},
  journal={arXiv preprint arXiv:2505.09388},
  year={2025},
  eprint={2505.09388},
  archivePrefix={arXiv},
  url={https://arxiv.org/abs/2505.09388},
}

@inproceedings{rag,
  title={Retrieval-Augmented Generation for Knowledge-Intensive {NLP} Tasks},
  author={Lewis, Patrick and Perez, Ethan and Piktus, Aleksandra and Petroni, Fabio and Karpukhin, Vladimir and Goyal, Naman and Kuettler, Heinrich and Lewis, Mike and Yih, Wen-tau and Rocktaschel, Tim and Riedel, Sebastian and Kiela, Douwe},
  booktitle={Advances in Neural Information Processing Systems},
  volume={33},
  year={2020},
  url={https://proceedings.neurips.cc/paper/2020/hash/6b493230205f780e1bc26945df7481e5-Abstract.html},
  pages={9459--9474},
}

@inproceedings{readonce,
  title={ReadOnce Transformers: Reusable Representations of Text for Transformers},
  author={Lin, Shih-Ting and Sabharwal, Ashish and Khot, Tushar},
  year={2021},
  booktitle={Proceedings of the 59th Annual Meeting of the Association for Computational Linguistics and the 11th International Joint Conference on Natural Language Processing (Volume 1: Long Papers)},
  url={https://aclanthology.org/2021.acl-long.554/},
  pages={7129--7141},
  doi={10.18653/v1/2021.acl-long.554},
}

@article{ruler,
  title={{RULER}: What's the Real Context Size of Your Long-Context Language Models?},
  author={Hsieh, Cheng-Ping and Sun, Simeng and Kriman, Samuel and Acharya, Shantanu and Rekesh, Dima and Jia, Fei and Zhang, Yang and Ginsburg, Boris},
  journal={arXiv preprint arXiv:2404.06654},
  year={2024},
  url={https://arxiv.org/abs/2404.06654},
}

@article{sempic,
  title={{SemPIC}: Learning Semantic Position-Independent {KV} Caches},
  author={Xie, Hui and Xiao, Peng and Deng, Yutong and Dou, Shuoran and Yang, Jian and Guo, Jinyang},
  year={2026},
  journal={arXiv preprint arXiv:2607.28069},
  url={https://arxiv.org/abs/2607.28069},
}

@inproceedings{turborag,
  title={{TurboRAG}: Accelerating Retrieval-Augmented Generation with Precomputed {KV} Caches for Chunked Text},
  author={Lu, Songshuo and Wang, Hua and Rong, Yutian and Chen, Zhi and Tang, Yaohua},
  year={2025},
  booktitle={Proceedings of the 2025 Conference on Empirical Methods in Natural Language Processing},
  pages={6588--6601},
  url={https://aclanthology.org/2025.emnlp-main.334/},
  doi={10.18653/v1/2025.emnlp-main.334},
}

@inproceedings{xccache,
  title={{XC-Cache}: Cross-Attending to Cached Context for Efficient {LLM} Inference},
  author={Monteiro, Jo{\~a}o and Marcotte, {\'E}tienne and No{\"e}l, Pierre-Andr{\'e} and Zantedeschi, Valentina and V{\'a}zquez, David and Chapados, Nicolas and Pal, Christopher and Taslakian, Perouz},
  year={2024},
  booktitle={Findings of the Association for Computational Linguistics: EMNLP 2024},
  url={https://aclanthology.org/2024.findings-emnlp.896/},
  pages={15284--15302},
  doi={10.18653/v1/2024.findings-emnlp.896},
}

@inproceedings{kivi,
 title={{KIVI}: A Tuning-Free Asymmetric 2bit Quantization for {KV} Cache},
 author={Liu, Zirui and Yuan, Jiayi and Jin, Hongye and Zhong, Shaochen and Xu, Zhaozhuo and Braverman, Vladimir and Chen, Beidi and Hu, Xia},
 booktitle={Proceedings of the 41st International Conference on Machine Learning},
 series={Proceedings of Machine Learning Research},
 volume={235}, pages={32332--32344}, publisher={PMLR}, year={2024},
 url={https://proceedings.mlr.press/v235/liu24bz.html}
}

@inproceedings{kvquant,
 title={{KVQuant}: Towards 10 Million Context Length {LLM} Inference with {KV} Cache Quantization},
 author={Hooper, Coleman and Kim, Sehoon and Mohammadzadeh, Hiva and Mahoney, Michael W. and Shao, Yakun Sophia and Keutzer, Kurt and Gholami, Amir},
 booktitle={Advances in Neural Information Processing Systems},
 volume={37}, pages={1270--1303}, publisher={Curran Associates, Inc.}, year={2024},
 doi={10.52202/079017-0040},
 url={https://proceedings.neurips.cc/paper_files/paper/2024/hash/028fcbcf85435d39a40c4d61b42c99a4-Abstract-Conference.html}
}

@inproceedings{smoothquant,
 title={{SmoothQuant}: Accurate and Efficient Post-Training Quantization for Large Language Models},
 author={Xiao, Guangxuan and Lin, Ji and Seznec, Mickael and Wu, Hao and Demouth, Julien and Han, Song},
 booktitle={Proceedings of the 40th International Conference on Machine Learning},
 series={Proceedings of Machine Learning Research},
 volume={202}, pages={38087--38099}, publisher={PMLR}, year={2023},
 url={https://proceedings.mlr.press/v202/xiao23c.html}
}

@misc{qwen35config,
  author       = {{Qwen Team}},
  title        = {{Qwen3.5-9B}: Official Model Configuration},
  howpublished = {Hugging Face model repository},
  year         = {2026},
  url          = {https://huggingface.co/Qwen/Qwen3.5-9B/blob/main/config.json},
  note         = {Accessed September 25, 2026}
}

@misc{qwen38config,
  author       = {{Qwen Team}},
  title        = {{Qwen3.8-27B}: Official Model Configuration},
  howpublished = {Hugging Face model repository},
  year         = {2026},
  url          = {https://huggingface.co/Qwen/Qwen3.8-27B/blob/main/config.json},
  note         = {Accessed September 25, 2026}
}

@inproceedings{cachegen,
  title={{CacheGen}: {KV} Cache Compression and Streaming for Fast Large Language Model Serving},
  author={Liu, Yuhan and Li, Hanchen and Cheng, Yihua and Ray, Siddhant and Huang, Yuyang and Zhang, Qizheng and Du, Kuntai and Yao, Jiayi and Lu, Shan and Ananthanarayanan, Ganesh and Maire, Michael and Hoffmann, Henry and Holtzman, Ari and Jiang, Junchen},
  booktitle={Proceedings of the ACM SIGCOMM 2024 Conference},
  year={2024},
  doi={10.1145/3651890.3672274},
  url={https://arxiv.org/abs/2310.07240}
}

@misc{terminalbench21,
  author       = {{Terminal-Bench Team}},
  title        = {{Terminal-Bench 2.1}},
  year         = {2026},
  howpublished = {Official benchmark and dataset repository},
  url          = {https://github.com/harbor-framework/terminal-bench-2-1},
  note         = {Accessed September 25, 2026}
}
\clearpage
\appendix
\section{Reproduction Details and Evidence Boundaries}
\label{app:protocol}
This appendix specifies protocols for the main-text experiments.

\subsection{Model, state, and training protocols}
The attention-only backbone is Qwen/Qwen3-8B, recorded revision
\texttt{b968826d9c46}, with 36 blocks, width 4096, 32 query heads,
eight KV heads of width 128, RoPE base $10^6$, and a configured window
of 40,960. Native diagnostics use BF16/SDPA; the complete precision
suite uses FP16 backbone computation with FP32 LoRA parameters.
H16 means the native 16-bit state format in its own experiment.
The custom precision reader is recorded as \texttt{final4000}, rank 32,
$\alpha=32$, trained without hidden-state quantization. H16/H8/H4 share its weights; identity with the native-reuse checkpoint is unverified.

The later documented training implementation uses 64 PG-19 books,
4,096-token windows (seven context chunks and one query chunk),
4,000 steps, AdamW with $\beta=(.9,.95)$, zero weight decay,
peak learning rate $10^{-4}$, 50 warmup steps, cosine decay,
clipping 1.0, seed 42, and gradient checkpointing. The frozen teacher
processes the full ordered window causally; only query positions
contribute to Equation~\ref{eq:distill}. Suffix Q/K/V/O and MLP
projections receive LoRA; the $j=12$ adapter has 58.20M parameters.
A separate archived CoMem manuscript lists $\alpha=64$,
2,048-token windows, and an eight-GPU training schedule.
These records do not establish cross-run checkpoint identity or training cost.

For H8/H4, groups are flattened within a chunk and padded as required.
Two BF16 metadata values contribute four bytes per 64 values;
reconstruction uses their stored rounded values, not the original FP32
parameters. Store counts unique backing after native chunk tensors are released,
excluding reconstruction workspace, weights, and request-local KV. Source, chunking, encoder, split, or Write-position changes invalidate
affected states; suffix-only adaptation does not. Overlap adds prefix dependencies.

\subsection{Input boundaries, selection, and generation}
The reusable serving and precision interfaces receive source and query
separately, fixing source chunking before questions arrive. Some native
quality drivers instead split a complete formatted prompt and take its last
chunk as the query; this may contain a document tail, and query-dependent
blocks must be rewritten. Their within-cohort controls share the boundary,
but those quality examples alone do not establish cross-query reuse.

Token-ID BM25 uses $k_1=1.5,b=.75$ and smoothed Robertson IDF,
without word normalization. Three rounds select up to four new chunks
per round, using newly retrieved chunks as the frontier; results are
deduplicated and restored to source order. Selection is fixed across
precision variants. Queries see all selected chunks in upper-layer causal
attention. The block-diagonal control removes only evidence-to-evidence
cross-chunk attention while preserving query access.

Precision quality uses sink 151643, EOS 151645, first-step EOS suppression,
and greedy generation that stops before a later EOS is appended.
The custom trainer's BOS fallback is 151645; this train/evaluation mismatch
is shared across precision variants. RULER caps are 48 tokens and 60 for
tracking; LongEval 16; BABILong 20; QA6 uses 128 for NarrativeQA/Qasper,
64 for MultiFieldQA-en, and 32 for HotpotQA/2WikiMQA/MuSiQue.
Capped outputs are kept untrimmed for native scoring. Fixed-cost requests
instead generate exactly 32 or 128 tokens without EOS stopping.

\subsection{Benchmark support and precision results}
RULER precision means average three tasks (single-2, multikey-1,
variable tracking) at 8k, 16k, 32k, 64k, and 128k, with 100 examples/cell.
Tracking uses fractional recall over five references; other needle tasks
use substring credit. Native paired RULER-B uses a separately sampled
single-3 cohort, and is never subtracted from the precision cohort.
LongEval averages first-number exact match over five lengths, 100 each.
LongBench QA6 averages six dataset token-F1 means over 1,150 questions
in 880 within-dataset document groups. BABILong averages official
answer scores over qa1/qa2/qa5 and 0k/1k/2k/4k/8k/16k/32k,
100 examples per cell. Full-source unextended stress rows are not
length-extended quality upper bounds.
Hybrid Dense and native LoCoMo full-source controls (harness label
\texttt{kvdirect}) recompute all layers without retrieval or LoRA; they
do not implement the cache manager of \citet{kvdirect}.

The 8B no-LoRA row in Table~\ref{tab:quality} is the precision report's
frozen reader, not the native-reuse no-adapter row. Its RULER/LongEval/
QA6/BABILong scores are 16.01/0.0/9.98/38.38; its semantic LoCoMo
score is 392/1,540, or 25.45. The H16/H8/H4 LoCoMo counts are
678/679/671 accepted answers, or 44.03/44.09/43.57.
The judge is the method-blind \texttt{gpt-6-astra/high} reference-binary
rubric. It has no bound dated backend snapshot or supplied reliability
estimate; authentication routes changed during acquisition.
The 446 category-5 refusal items remain a separate diagnostic.
Native LoCoMo uses GPT-4o on categories 1--4 plus local abstention on
category 5, totaling 1,986 items across ten conversations.

The hybrid extension records $j=6$ for Qwen3.5-9B and $j=21$ for
Qwen3.8-27B, native chat templates with thinking disabled, and separate
4,000-step readers. Its H16 full LoCoMo protocol uses Astra/low plus
local abstention over 1,986 items. Hybrid low-bit aggregates lack independently supplied checkpoint identity,
per-item predictions, inference dtype, judge binding, and confidence intervals.
They remain descriptive; 8B supplies the principal controlled precision evidence.

\subsection{Terminal-Bench 2.1 agent evaluation}
The additional agent study uses the 89-task Terminal-Bench 2.1 manifest
\citep{terminalbench21}. Each task runs in its own container and its
task-specific verifier supplies a binary reward. We retain one
verified outcome per task and method. Dense uses full-context Qwen3.8-27B
in a vLLM serving path without the EncBank suffix adapter. EncBank-H16
uses the same backbone family, $j=21$, the adapted suffix, disjoint
512-token chunks, and up to 12 selected chunks ($k=12$) in a Transformers
serving path. Each task owns a distinct session. The native chat template
tokenizes its accumulated instructions, chat messages, and tool output.
The fixed initial prefix is kept separately; previous exchanges are
partitioned into 512-token chunks and keyed by token identity. The H21
bank reuses unchanged chunks and independently encodes new ones with the
lower 21 blocks. For a new call, token-ID BM25 uses the initial task
question and latest user message to select up to $k$ archived chunks;
they are reordered chronologically and joined with the fixed prefix and
current exchange. The suffix reader with LoRA attends causally over this
pack and generates the next agent response. The session persists across
the task's model calls, then its H bank is released at task closure.
The agent/controller paths are custom and changed
during infrastructure recovery. Hence these are not matched-backend,
matched-adapter ablations or official leaderboard submissions.

Table~\ref{tab:terminalbench} reports complete 89-task outcomes for Dense
and EncBank-H16 at $k=12$, with success rates computed over the same
full-task denominator. A valid outcome requires the task verifier result
and trial-closure evidence; earlier valid scores are retained across
continuation jobs, without counting a task twice. Infrastructure
interruption, incomplete replies, and unstarted tasks are not counted as
verifier failures, whereas reaching the model's native 262,144-position
capacity is a recorded failure with score zero. No user-imposed generation,
request, or task-time cutoff applies to these continuation runs. The
official leaderboard specifies at least five trials per task and public
validation; this single-retained-outcome study is not directly comparable
to it. Aggregate provenance, including the source of the completed
EncBank result, is recorded in the source package.

\subsection{Uncertainty, clean filtering, and timing}
Reported intervals preserve the original analyses, not newly inferred
ones. FP16 QA contrasts use 10,000 paired exact-document-cluster draws,
item-weighted means within each dataset, and seed 20260911
(NarrativeQA: 20260912). The MuSiQue interval is unadjusted for
multiple comparisons. Distillation validation uses 10,000 paired
book-level resamples, keeping two windows/book together.
Clean LongBench filtering uses normalized 13-gram containment below .10
against the full PG-19 training split, removing 96 NarrativeQA and
one 2WikiMQA example. Its paired intervals resample source-document
clusters within datasets, then average six means (10,000 draws,
seed 20260913).

The latency-calibrated budget control chooses $k$ on three reserved
RULER multikey/32k documents, indices 900--902, disjoint from quality
indices 0--99. One H20, three processes, five warmups and 20 repetitions
per budget give BM25 replay/H16 TTFT 703.7/698.3 ms (GPU store),
within the declared $\pm5\%$ band. Quality comprises nine 100-example
cells: BABILong qa1/qa2 at 4k/16k, LongEval and multikey at 8k/16k,
and the first 100 LoCoMo questions from conversation zero.
Hierarchical intervals use 100,000 paired resamples of cells and then
examples within selected cells, seed 20260804. No conversation-level
inference is available for that one-conversation slice.

The local native systems environment is BF16/SDPA on RTX 5090,
with model weights resident, PyTorch 2.7.1+cu128 and Transformers 5.16.1.
Full-source costs use three processes and three saved documents/length;
phases are timed separately, and OOM is an actual failed attempt.
Matched chunk-KV costs use three processes, three formal repetitions,
and one warmup for each method/adapter/example. The 32k slice in
Table~\ref{tab:kv} is part of a 20-example, 1,080-trial cost study;
its 500-example quality study produces 3,000 predictions.
Cold E2E is measured continuously through token 128, not formed by
adding independently measured medians.

The FP16 precision timing study uses native SDPA and tokenwise query
ingestion on RTX 5090. The methods were acquired in nonrandom order
on different days with differing execution graphs. Sampled device
baselines are 2.470 GiB for Dense/H and 2.472 GiB for KIVI.
A fixed set of six RULER/8k documents provides six warmups and 18
measured fresh Write/Read pairs per method with exactly 32 output tokens.
Warm TTFT excludes document Write but includes query ingestion,
selection/reconstruction, and first-token production. Store counts
unique persistent GPU backing after native chunk tensors are released.
A 250-ms global-device sample can include background changes;
it differs from allocator peaks in the native scaling test.
Resident occupancy levels run sequentially with no eviction/offload;
only the first six known targets are queried. The native amortization
grid includes fetch, Read and decode but cancels common selection.
For 32k CPU-pinned storage, preparation is 2.253 s versus 2.16 ms
for the raw-token index. Its assumptions exclude corpus edits,
cache misses, concurrent tenants, and eviction.

\subsection*{Ethics and reproducibility}
Residual states require source-equivalent access, isolation, and deletion.
No new human-subject data are collected; external assets retain their access
and redistribution terms. The package includes provenance and build/plotting
commands. LLM tools assisted drafting; authors are responsible for all claims.

\end{document}